\documentclass[
]{ceurart}

\usepackage{adjustbox}
\usepackage{xcolor}
\usepackage{caption}
\usepackage{subcaption}

\begin{document}

\copyrightyear{2021}
\copyrightclause{Copyright for this paper by its authors.
  Use permitted under Creative Commons License Attribution 4.0
  International (CC BY 4.0).}

\conference{NeSy'20/21 @ IJCLR: 15th International Workshop on
Neural-Symbolic Learning and Reasoning, October 25--27, 2021}

\title{An Insect-Inspired Randomly, Weighted Neural Network with Random Fourier Features For Neuro-Symbolic Relational Learning}

\author[1]{Jinyung Hong}[%
orcid=0000-0003-4429-3311,
email=jhong53@asu.edu,
url=https://www.linkedin.com/in/jyhong0304/,
]

\author[1,2,3,4]{Theodore P.~Pavlic}[%
orcid=0000-0002-7073-6932,
email=tpavlic@asu.edu,
url=https://isearch.asu.edu/profile/1995237,
]

\address[1]{School of Computing and Augmented Intelligence, Arizona State University, Tempe, AZ 85281, USA}

\address[2]{School of Sustainability, Arizona State University, Tempe, AZ 85281, USA}

\address[3]{School of Complex Adaptive Systems, Arizona State University, Tempe, AZ 85281, USA}

\address[4]{School of Life Sciences, Arizona State University, Tempe, AZ 85281, USA}

\begin{abstract}
    The computer-science field of Knowledge Representation and Reasoning~(KRR) aims to understand, reason, and interpret knowledge as efficiently as human beings do. Because many logical formalisms and reasoning methods in the area have shown the capability of higher-order learning, such as abstract concept learning, integrating artificial neural networks~(ANNs) with KRR methods for learning complex and practical tasks has received much attention. For example, Neural Tensor Networks~(NTNs) are neural-network models capable of transforming symbolic representations into vector spaces where reasoning can be performed through matrix computation; when used in Logic Tensor Networks~(LTNs), they are able to embed first-order logic symbols such as constants, facts, and rules into real-valued tensors.
The integration of KRR and ANN suggests a potential avenue for bringing biological inspiration from neuroscience into KRR.
However, higher-order learning is not exclusive to human brains. Insects, such as fruit flies and honey bees, can solve simple associative learning tasks and learn abstract concepts such as ``sameness'' and ``difference,'' which is viewed as a higher-order cognitive function and typically thought to depend on top-down neocortical processing. 
Empirical research with fruit flies strongly supports that a randomized
representational architecture is used in olfactory processing in insect brains.
Based on these results, we propose a Randomly Weighted Feature Network~(RWFN)
that incorporates randomly drawn, untrained weights in a encoder that uses an
adapted linear model as a decoder. The randomized projections between input
neurons and higher-order processing centers in the input brain is mimicked in
RWFN by a single-hidden-layer neural network that specially structures latent
representations in the hidden layer using random Fourier features that better
represent complex relationships between inputs using kernel approximation.
Because of this special representation, RWFNs can effectively learn the degree of relationship among inputs by training only a linear decoder model.
We compare the performance of RWFNs to LTNs for Semantic Image Interpretation~(SII) tasks that have been used as a representative example of how LTNs utilize reasoning over first-order logic to surpass the performance of solely data-driven methods. We demonstrate that compared to LTNs, RWFNs can achieve better or similar performance for both object classification and detection of the \emph{part-of} relations between objects in SII tasks while using much far fewer learnable parameters~(1:62 ratio) and a faster learning process~(1:2 ratio of running speed). Furthermore, we show that because the randomized weights do not depend on the data, several decoders can share a single randomized encoder, giving RWFNs a unique economy of spatial scale for simultaneous classification tasks.
\end{abstract}


\maketitle

\section{Introduction}
\label{sec:intro}
The human brain has an extraordinary ability to memorize and learn new things to solve a variety of problems with difficulty ranging from trivial to complex. To understand the cognitive architecture of the brain, research on producing a wiring diagram of the connections among all neurons, called \emph{Connectomics}~\citep{seung2012connectome}, has focused not only on the human brain~\citep{sporns2005human} but also on the brains of insects~\citep{eichler2017complete, takemura2017connectome}, and such research has influenced the development of machine learning and Artificial Intelligence~(AI)~\citep{hassabis2017neuroscience}. However, far more is known about the function of coarse-grained, high-level structures in the brain than the neuron-scale layout of important brain regions. Similarly, the high degrees of freedom in artificial neural networks~(ANN) has provided an opportunity for the introduction of Knowledge Representation and Reasoning~(KRR) to constructively constrain ANN architectures and training methods. In particular, combining KRR techniques with ANNs promises to enhance the high performance of modern AI with explainability and interpretability, which is necessary for generalized human insight and increased trustworthiness.

Several recent studies across statistical relational learning~(SRL),
neural-symbolic computing, knowledge completion, and approximate
inference~\citep{koller2007introduction, garcez2008neural,
pearl2014probabilistic, nickel2015review} have shown that neural networks can
be integrated with logical systems to perform robust learning and effective
inference while also providing increased interpretability from symbolic
knowledge extraction. These neural-network knowledge representation approaches
use \emph{relational embedding} to represents relational predicates in a neural
network~\citep{sutskever2009using, bordes2011learning, socher2013reasoning,
santoro2017simple}. For example, Neural Tensor Networks~(NTNs) are structured
to encode the degree of association among pairs of entities in the form of
tensor operations on real-valued vectors~\citep{socher2013reasoning}. These
NTNs have been synthesized with neural symbolic
integration~\citep{garcez2008neural} in the development of Logic Tensor
Networks~(LTNs)~\citep{serafini2016logic}, which can extend the power of NTNs
to reason over first-order many-valued logic~\citep{bergmann2008introduction}.

Although KRR aims to lift the reasoning ability of computers to that of humans,
such higher-order learning and reasoning capabilities are not unique to humans.
Insect neuroscience has shown that insects show sophisticated and complex
behaviors even though they possess miniature central nervous systems compared
to the human brain~\citep{avargues2011visual}. For example, attention-like
processes have been demonstrated in in fruit flies and honey
bees~\citep{van2003salience, spaethe2006honeybees}, and concept learning has
been shown in bees~\citep{avargues2012simultaneous}. Specifically, it has been
shown that the honey bee brain contains high levels of cognitive sophistication
so that it can learn relational concepts such as ``same,'' ``different,''
``larger than,'' ``better than,'' among others, and researchers continue to
study the neurobiological mechanisms and computational models underlying these
capabilities~\citep{avargues2013conceptual}. Just as KRR is now being used to
better shape ANN's for more sophisticated reasoning and increased
interpretability, the architectures demonstrated in the honey bee brain may
provide insights into how to augment ANN's with higher-order reasoning
abilities akin to those demonstrated in insects.

In this paper, we propose Randomly Weighted Feature Networks~(RWFNs), an
insect-brain-inspired single-hidden-layer neural network for relational
embedding that incorporates randomly drawn, untrained weights in its
encoder with a trained linear model as a decoder. Our approach is mainly
motivated by neural circuits in the insect brain centered around the
Mushroom Body~(MB). The MB, analogous to the neocortex in humans, is a vital
region of the insect brain supporting concept learning because it is
responsible for stimulus identification, categorization, and element
learning~\citep{menzel2001searching, galizia2014olfactory,
bazhenov2013computational}. We can model the MB as a neural network
model with three layers: Input Neurons~(INs)~-- Kenyon Cells~(KCs)~--
mushroom body Extrinsic Neurons~(ENs). One of the remarkable properties
of MB is that the connections between INs and KCs are relatively random
and sparse~\citep{caron2013random}. To mimic this characteristic, we used a
random weight matrix to transform the input between the input and hidden layers
to generate the latent representation of the relationship between real-valued
input entities. By doing so, the learning process involves only the training of
the weights between the hidden and output layers, which is simple and fast. In
contrast, a conventional LTN would incorporate an NTN specially trained to
capture logical relationships present in data, which requires more learning
parameters and a more complex learning process.

Our method is also influenced by random Fourier
features~\citep{rahimi2007random}, a kernel approximation method that overcomes
the issues of conventional kernel machines or kernel
methods~\citep{smola1998learning}. Kernel methods are one of the most powerful
and theoretically grounded approaches for nonlinear statistical learning
problems, including classification, regression, clustering,
and others~\citep{zhu2005kernel, drucker1997support, dhillon2004kernel}.
However, the main issue of the kernel method is the lack of scalability for
large datasets and a slow training process~\citep{rahimi2007random,
liu2020random}. Random Fourier features can address these issues by
approximating the kernel function by using an explicit feature mapping that
projects the input data into a randomized feature space and by applying faster
linear models to learn. Interestingly, the random Fourier features model can
also be viewed as a class of a single-hidden-layer neural network model with a
fixed weight between the input and hidden layers. Thus, we leveraged this to
substitute the tensor operations in conventional NTNs that model the linear
interactions between entities and utilized the projection from input into
another space as another feature representation in the hidden layer of our
model to learn relationships.

Thus, our proposed model is an insect-inspired
single-hidden-layer network with latent representation derived by the
integration of the input transformation between INs and KCs and random
Fourier features, and it only requires training of a linear decoder. By
applying the model to solve the Semantic Image Interpretation~(SII)
tasks, we show that a trained linear decoder in RWFNs can effectively
capture the likelihood of \emph{part-of} relationships at a level of
performance exceeding that of traditional LTNs, even with far fewer
parameters and a faster learning process. To the best of our knowledge,
this is the first research to integrate both insect neuroscience and
neuro-symbolic approaches for reasoning under uncertainty and for
learning in the presence of data and rich knowledge. Furthermore,
because the encoder weights in our model do not depend upon the data,
the single encoder can be shared among several decoders, each trained
for a different classifier, giving RWFNs an economy of spatial scale in
our model applications where several classifiers need to be used
simultaneously.

\section{Related Work and Background}
\label{sec:related}

\paragraph{Insect Neuroscience}
The MB in the insect brain receives processed olfactory, visual, and
mechanosensory stimuli~\citep{mobbs1982brain} and is viewed as the critical
region responsible for multimodal associative
learning~\citep{menzel2001searching}. In the fruit fly, thousands of Kenyon
Cells~(KCs) in the MB each receive a set of random $\sim$7 inputs from
INs~\citep{caron2013random, inada2017origins}, and this is similarly true for
honey bees~\citep{peng2017simple}. A simplified neural circuit modeling the MB
is a neural network with three layers consisting of: i) INs that provide
olfactory, visual, and mechanosensory inputs, ii) KCs generating the
sparse-encoding of sensory stimuli, and iii) ENs for activating several
different behavioral responses~\citep{cope2018abstract}. In particular, INs
receive various inputs from Antennal Lobe~(AL) glomeruli, Medulla, and Lobula
optic neuropils~\citep{cope2018abstract}. For simplicity, we focus on the
olfactory pathway between glomeruli in the AL and KCs in the
MB~\citep{cope2018abstract, peng2017simple}. The insect olfactory neural
circuit has a divergence--convergence structure where $\sim$800 AL glomeruli
form a coded feature vector that expands into a spare representation across
$\sim$170,000 KCs, and these are decoded by
$\sim$400 ENs that actuate motor pathways based on this
information-processing pipeline~\citep{peng2017simple}. This general
divergence--convergence structure applies equally as well to honey bees
and fruit flies~\citep{endo2020synthesis}; therefore, for modeling the
architecture of our methods for brevity, we interchangeably leverage the
neural circuits of olfactory nervous systems between these insects.

In this paper, the input transformation of odorant representation in the
AL to the higher-order representation across the KCs in the MB is shown
in the projection between the input and hidden layers of our model, and
this representation plays the critical role in learning the
relationships presented on the input.


\paragraph{Random Fourier Features}
Kernel machines, e.g., Support Vector
Machines~(SVMs)~\citep{boser1992training, cortes1995support}, have
received significant attention due to their capability for function
approximation and excellent performance of detecting decision boundaries
with enough training data. These methods use transformations, as
with a lifting function $\phi$, that help to better discriminate among
different inputs. Given dataset vector inputs $\textbf{x}, \textbf{y}
\in \mathbb{R}^d$, the kernel function $k(\textbf{x}, \textbf{y}) =
\langle\phi(\textbf{x}), \phi(\textbf{y})\rangle$ represents the
similarity (i.e., inner product) between $\textbf{x}$ and $\textbf{y}$
in the transformed space. However, because of the potential complexity
of the transformation $\phi$, learning the kernel function $k$ may require
significant computational and storage costs.

Random Fourier features~\citep{rahimi2007random}, instead, provide a data
transformation that permits using a far less expensive approximation of
the kernel function. For each vector input $\textbf{x} \in
\mathbb{R}^d$, the technique applies a randomized feature function
$\textbf{z}: \mathbb{R}^d \to \mathbb{R}^D$ (generally, $D \gg d$ with
sample size $N \gg D$) that maps $\textbf{x}$ to evaluations of $D$
random Fourier basis from the Fourier transform of kernel $k$. In this
transformed space, kernel evaluations can be approximated by linear
operations, as in:
\begin{equation}
    k(\textbf{x}, \textbf{y})
        = \langle \phi(\textbf{x}), \phi(\textbf{y}) \rangle
        \approx \textbf{z}(\textbf{x})^{\top}\textbf{z}(\textbf{y})
    \label{eq:random_fourier_feature}
\end{equation}
Thus, by transforming the input with $\textbf{z}$, fast linear learning
methods can be leveraged to approximate the evaluations of nonlinear
kernel machines.

In this paper, we use random Fourier features as latent representations
that reduce the complexity of learning relations among real-valued entities. As
described in Section~\ref{sec:intro}, adaptable NTNs within LTNs have been used
to encode relationships among real-valued entities. We replace the adaptable
NTNs with random Fourier features that have high expressiveness with low
decoding overhead. Details and intuitions will be given in
Section~\ref{sec:RWFNs}.

\paragraph{Logic Tensor Networks~(LTNs)}
The RWFNs we propose are meant to improve upon LTNs for statistical relational
learning tasks. LTNs integrate learning based on
NTNs~\citep{socher2013reasoning} with reasoning using first-order, many-valued
logic~\citep{bergmann2008introduction}, all implemented in
TensorFlow~\citep{serafini2016logic}. Here, we briefly introduce LTN syntax and
semantics for use in mapping logical symbols to numerical values and
learning reasoning relations among real-valued vectors using the logical
formulas.

Although a first-order-logic~(FOL) language $\mathcal{L}$ and its
signature are defined by consisting of three disjoint sets~-- i)
$\mathcal{C}$ (constants), ii) $\mathcal{F}$ (functions) and iii)
$\mathcal{P}$ (predicate)~-- we ignore function symbols $\mathcal{F}$
because they are not used in SII tasks that we focus on here. For any
predicate symbol $s$, $\alpha(s)$ can be described as its \emph{arity},
and logical formulas in $\mathcal{L}$ enable the description of
relational knowledge. The objects being reasoned over with FOL are
mapped to an interpretation domain $\subseteq \mathbb{R}^{n}$ so that
every object is associated with an $n$-dimensional vector of real
numbers. Intuitively, this $n$-tuple indicates $n$ numerical features of
an object. Thus, predicates are interpreted as fuzzy relations on real
vectors. With this numerical background, we can now define the numerical
\emph{grounding} of FOL with the following semantics; this grounding is
necessary for NTNs to reason over logical statements.

Let $n \in \mathbb{N}$. An $n$-grounding, or simply grounding,
$\mathcal{G}$ for a FOL $\mathcal{L}$ is a function defined on the
signature of $\mathcal{L}$ satisfying the following conditions:
\begin{gather*}
    \mathcal{G}(c) \in \mathbb{R}^n \; \text{for every constant symbol} \; c
        \in \mathcal{C}\\
    \mathcal{G}(P) \in \mathbb{R}^{n \cdot \alpha(f)} \to [0,1]
        \; \text{ for predicate sym.~} \; P \in \mathcal{P}
\end{gather*}
Given a grounding $\mathcal{G}$, the semantics of closed terms and
atomic formulas is defined as follows:
\begin{gather*}
    \mathcal{G}(P(t_1, \dots, t_m)) \triangleq \mathcal{G}(P)(\mathcal{G}(t_1),
        \dots, \mathcal{G}(t_m))
\end{gather*}
The semantics for connectives, such as $\mathcal{G}(\neg \phi),
\mathcal{G}(\phi \land \psi), \mathcal{G}(\phi \lor \psi)$, and
$\mathcal{G}(\phi \rightarrow \psi)$, can be computed by following the
fuzzy logic such as the Lukasiewicz
$t$-norm~\citep{bergmann2008introduction}.

A partial grounding $\hat{\mathcal{G}}$ can be defined on a subset of
the signature of $\mathcal{L}$. A grounding $\mathcal{G}$ is said to be
a completion of $\hat{\mathcal{G}}$ if $\mathcal{G}$ is a grounding for
$\mathcal{L}$ and coincides with $\hat{\mathcal{G}}$ on the symbols
where $\hat{\mathcal{G}}$ is defined. Let \emph{GT} be a grounded theory
which is a pair $\langle \mathcal{K}, \hat{\mathcal{G}}\rangle$ with a
set $\mathcal{K}$ of closed formulas and a partial grounding
$\hat{\mathcal{G}}$. A grounding $\mathcal{G}$ satisfies a \emph{GT}
$\langle \mathcal{K}, \hat{\mathcal{G}}\rangle$ if $\mathcal{G}$
completes $\hat{\mathcal{G}}$ and $\mathcal{G}(\phi)=1$ for all $\phi
\in \mathcal{K}$. A \emph{GT} $\langle \mathcal{K},
\hat{\mathcal{G}}\rangle$ is satisfiable if there exists a grounding
$\mathcal{G}$ that satisfies $\langle \mathcal{K},
\hat{\mathcal{G}}\rangle$. In other words, deciding the satisfiability
of $\langle \mathcal{K}, \hat{\mathcal{G}}\rangle$ amounts to searching
for a grounding $\mathcal{G}$ such that all the formulas of
$\mathcal{K}$ are mapped to 1. If a \emph{GT} is not satisfiable, the
best possible satisfaction that we can reach with a grounding is of our
interest.

Grounding $\mathcal{G}^*$ captures the implicit correlation between
quantitative features of objects and their categorical/relational
properties. The grounding of an $m$-ary predicate $P$, namely
$\mathcal{G}(P)$, is defined as a generalization of the
NTN~\citep{socher2013reasoning}, as a function from $\mathbb{R}^{mn}$ to
$[0,1]$, as follows:
\begin{equation}
    \label{eq:ltn_predicate}
    \mathcal{G}_{LTN}(P)(\textbf{v})
    = \sigma (u_{P}^{\top} \texttt{f} (\textbf{v}^{\top} W_{P}^{[1:k]} \textbf{v} + V_{P} \textbf{v} + b_P))
\end{equation}
where $\textbf{v} = \langle \textbf{v}_1^\top, \dots, \textbf{v}_m^\top
\rangle^\top$ is the $mn$-ary vector obtained by concatenating each
$\textbf{\texttt{v}}_i$. $\sigma$ is the sigmoid function and
$\texttt{f}$ is the hyperbolic tangent ($\tanh$). The parameters for $P$
are: $W_{P}^{[1:k]}$, a 3-D tensor in $\mathbb{R}^{k \times mn \times
mn}, V_{P} \in \mathbb{R}^{k \times mn}, b_P \in \mathbb{R}^{k}$ and
$u_P \in \mathbb{R}^k$. Because our RWFN model can be used to ground a
predicate as $\mathcal{G}_{RWFN}(P)$, we can directly compare the
performance of RWFNs for the SII tasks with LTNs.

\section{Randomly Weighted Feature Networks (RWFNs)}
\label{sec:RWFNs}

In this section, we introduce the details of Randomly Weighted Feature Networks~(RWFNs). The underlying intuition behind the development of this model can be found in Appendix~\ref{app:intuition}.

\subsection{Model Architecture}
\begin{figure*}[t!]\centering%
    \begin{subfigure}[t]{0.49\textwidth}\centering%
        \includegraphics[width=\textwidth]{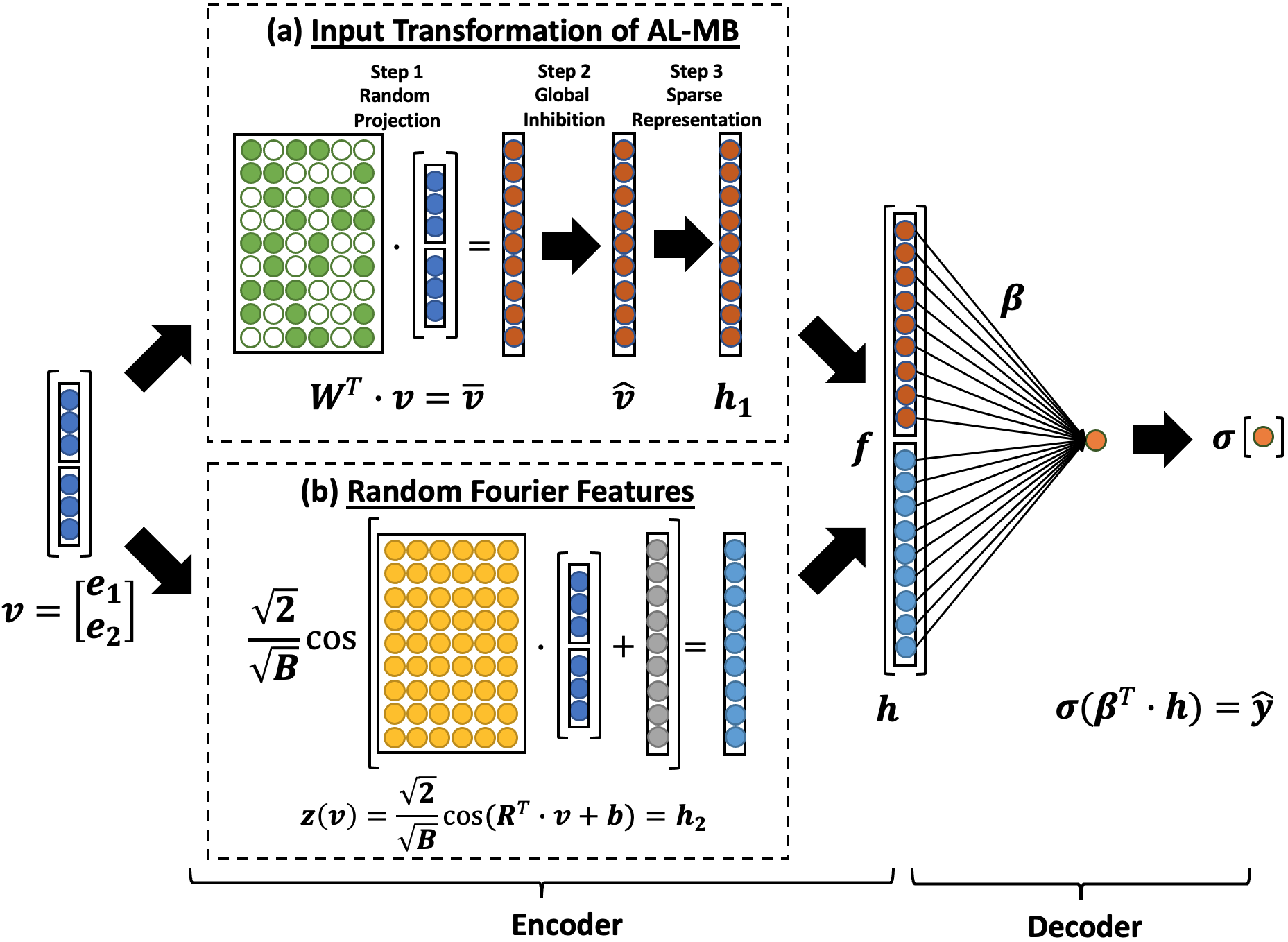}
        \caption{Visualization of the structure of the Randomly Weighted
    Feature Network. In the depicted case, the input vector $\textbf{v}$
    constitutes of two entities, $e_{1}, e_{2} \in \mathbb{R}^{3}$ and it
    shows to learn a binary relation between them $(e_{1}, R, e_{2})$, such
    as (Cat, hasPart, Tail).}
        \label{fig:rwfn-structure}
    \end{subfigure}
    \hfill
    \begin{subfigure}[t]{0.49\textwidth}\centering%
        \includegraphics[width=\textwidth]{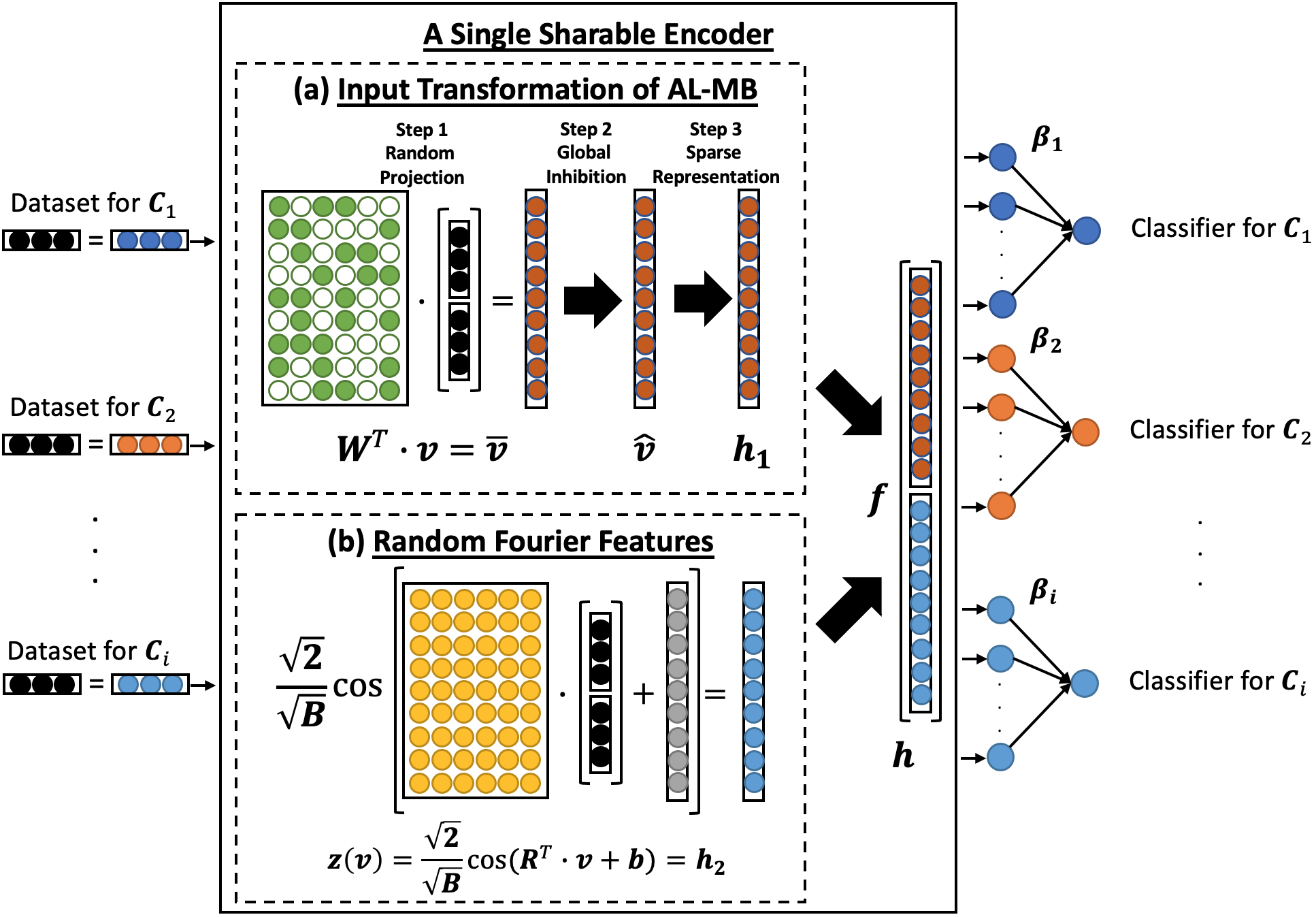}
        \caption{Visualization of the structure of the Randomly Weighted Feature
    Network with weight sharing. In the case of learning each classifier
    from the class $\mathcal{C}_{1}$ to the class $\mathcal{C}_{i}$, RWFNs
    allow us to use the same encoder to extract features from each data from the class $\mathcal{C}_{1}$ to the class $\mathcal{C}_{i}$.}
        \label{fig:rwfn_ws_structure}
    \end{subfigure}
    \hfill
    \caption{The architectures of RWFNs and RWFNs with weight sharing}
    \label{fig:rwfn_rwfn-ws}
\end{figure*}
%
Let the input vector $\textbf{v}$ be $[\textbf{v}_1^{\top}, \dots,
\textbf{v}_m^{\top}]^{\top}$, the $mn$-ary vector where $m$ is arity and
$n$ is the input dimension. We first define the two kinds of latent
representations: i) the input transformation between AL glomeruli and
KCs inspired by the insect brain, and ii) the transformed input using a
randomized feature mapping $\textbf{z}(\cdot)$ in random Fourier
features.

For the bio-inspired representation, we select $N_{in} \in [1,mn)$
indices of the input at random without replacement for each hidden
node\footnote{$N_{in} < mn$ to prevent hidden-unit outputs becoming
trivially 0 by Eq.~(\ref{eq:kcnet_intermediate}). In our setting,
$N_{in}=7$.}. In other words, the output $\Bar{v}_j$ of each hidden node
$j \in \{1,...,B\}$ is a weighted combination of all $mn$ inputs where
only $N_{in} < mn$ inputs have $w_{j,i}=1$ and all other inputs have
$w_{j,i}=0$. So the inputs are effectively gated by the weights on each
hidden node and the weight matrix $\textbf{W} \in \mathbb{R}^{mn \times
B}$ in this computation is random, binary, and sparse.

Once the $j$th hidden node has produced weighted sum $\Bar{v}_{j}$, the
post-processing step to produce intermediate output is performed in the
hidden layer. Mimicking Eq.~(\ref{eq:r_out_bio}) with $C=1$, the $j$th
intermediate output $\hat{x}_{j}$ is:
\begin{equation}
    \hat{v}_{j} = \Bar{v}_{j} - \mu, \;\; \mu = \frac{1}{B}\sum_{i=1}^{B} \Bar{v}_{i}
    \label{eq:kcnet_intermediate}
\end{equation}
where $B$ is the number of hidden units. Therefore, the sparse output of
the $j$-th KC node can be defined as
    $h_{j}^{(1)} = g(\hat{v}_{j})$
where $g$ is the ReLU function~\citep{glorot2011deep} that allows the
model to produce sparse hidden output, which is more biologically
plausible. By doing so, we define the output vector as $\textbf{h}_{1} =
[h_{1}^{(1)}, \dots, h_{B}^{(1)}]^{\top}$. 

On the other hand, to generate random Fourier features, we used a
randomized feature function $\textbf{z}(\cdot)$ in
\citep{rahimi2007random, sutherland2015error}, we can project the input
as follows:
\begin{equation}
    \textbf{h}_{2} = \textbf{z}(\textbf{v}) = \frac{\sqrt{2}}{\sqrt{B}}\cos(\textbf{R}^{\top}\textbf{v} + \textbf{b})
\label{eq:rwfn_second_rep}
\end{equation}
where $\textbf{R} \sim Normal^{mn \times B}(0,1)$ and $\textbf{b} \sim
Uniform^{B}(0, 2\pi)$, which is Gaussian kernel approximation.
Consequently, the output vector $\textbf{h}_{2}$ can be considered as another latent representation of relationship among input.
\Citet{rahimi2007random},
\citet{sutherland2015error}, and \citet{liu2020random} provide theoretical
derivations of kernel approximation and comparative analyses of
various kinds of random Fourier features.

Finally, using the above two latent representations, our RWFNs can be
defined as a function from $\mathbb{R}^{mn}$ to $[0,1]$:
\begin{equation}
    \mathcal{G}_{RWFN}(P)(\textbf{\texttt{v}})
    = \sigma \left(
        \boldsymbol\beta^\top
        \textbf{h}
    \right) \\
    = \sigma \left(
        \boldsymbol\beta^\top \texttt{f}\left(
            \begin{bmatrix}
                \textbf{h}_{1} \\
                \textbf{h}_{2}
            \end{bmatrix}
        \right)
    \right)
    \label{eq:rwfn}
\end{equation}
where $\textbf{h}$ is the final hidden representation obtained by applying the hyperbolic tangent~($\tanh$) function $\texttt{f}$  to the concatenation of $\textbf{h}_{1}$ and $\textbf{h}_{2}$, and $\sigma$ is the sigmoid function; the
$\tanh$ function was used for the numeric stabilization.
Because our model requires to adapt only $\boldsymbol\beta \in
\mathbb{R}^{2B}$, it possess a faster learning process with fewer
parameters compared to LTNs. Fig.~\ref{fig:rwfn-structure} shows a
visualization of the structure of our model.

\subsection{RWFNs with Weight Sharing}

In the insect brain, extrinsic neurons from the MB are
processed by several small, downstream neuropils that ultimately lead to
decision-making outcomes, such as muscle actuation. If we view these
small neuropils as decoding the complex representations in the MB, then different decoders responsible for different decisions all use information sourced from the same randomized representations in the MB. The MB can be viewed as a generalized encoder that is not tailored for a particular task; consequently, it provides a
shared resource to reduce the complexity of these downstream neuropils.

Because the weights of the randomized encoder of an RWFN are independent
of the training data, they can also serve as a shared resource for
multiple relatively simple (i.e., linear) downstream decoders trained
for different classifiers. We refer to this property as \emph{weight
sharing}. Fig.~\ref{fig:rwfn_ws_structure} shows a visualization of the
structure of our model applied with weight sharing to the learning of
$i$ different classifiers. The large solid box surrounds a single
encoder that serves as a common feature extractor for all classifiers.

The entities in the original definition of RWFNs in
Fig.~\ref{fig:rwfn-structure} become the placeholder to be injected by
the input for each classifier. Instead of generating the randomized
encoder for each classifier, each classifier uses the same encoder, and
training only requires learning the weights of that classifier's highly
simple linear decoder. This approach increases reusability and cost
efficiency in a way beyond what is possible with LTNs, which must train
all encoder and decoder networks separately for each classifier.

\section{Experimental Evaluation}
\label{sec:experiment}

To evaluate the performance of our proposed RWFNs over LTNs, we employ
both for SII tasks, which extract structured semantic descriptions from
images. Very few SRL applications have been applied to SII tasks because
of the high complexity involved with image learning.
\Citet{donadello2017logic} define two main tasks of SII as:~(i) the
classification of bounding boxes, and~(ii) the detection of the
\emph{part-of} relation between any two bounding boxes. They
demonstrated that LTNs can successfully improve the performance of
solely data-driven approaches, including the state-of-the art Fast
Region-based Convolutional Neural Networks~(Fast
R-CNN)~\citep{girshick2015fast}.
Our experiments are conducted by comparing the performance of two tasks
of SII between RWFNs and LTNs. These tasks are well defined in
first-order logic, and the codes implemented in TensorFlow framework
have been provided and can be used to compare the performance of LTNs
with RWFNs.

\subsection{Methods}

Here, we provide details of our experimental comparison of RWFNs and LTNs. We
utilize the formalization of SII in first-order logic from
\citet{donadello2017logic}. For brevity, we describe: (i) the difference
of the ground theories between RWFNs and LTNs, (ii) the data set used in
the experiments~(Appendix~\ref{app:details_exp}), and (iii) the RWFN and LTN hyperparameters used in the experiments~(Appendix~\ref{app:details_exp}). We omit other formalization details of the SII tasks that can be found elsewhere~\citep{donadello2017logic}.

\paragraph{Defining the Grounded Theories for RWFNs and LTNs} \label{sec:groundingRWTNs}

A set of bounding boxes of images correctly labelled with the classes
that they belong to and pairs of bounding boxes that properly labelled
with the \emph{part-of} relation were provided. These datasets can be
considered as a training set, and a grounded theory
$\mathcal{T}_{\text{LTN}} \triangleq \langle \mathcal{K},
\hat{\mathcal{G}}_{LTN} \rangle$ can be constructed. In particular,
$\mathcal{K}$ contains: (i) the set of closed literals $C_i(b)$ and $\texttt{partOf}(b, b')$ for every bounding box $b$
labelled with $C_i$ and for every pair of bounding boxes
$\langle b, b' \rangle$ connected by the \texttt{partOf}
relation, and~(ii) the set of the mereological constraints for
the \emph{part-of} relation, including asymmetric constraints,
lists of several parts of an object, or restrictions that whole
objects cannot be part of other objects and every part object
cannot be divided further into parts. 
Furthermore, the partial grounding $\hat{\mathcal{G}}_{LTN}$ is defined on all
bounding boxes of all the images in the training set where both
$class(C_i,b)$ and the bounding box coordinates are computed by
the Fast R-CNN object detector. $\hat{\mathcal{G}}$ is not
defined for the predicate symbols in $\mathcal{P}$ and is to be
learned.

A grounded theory $\mathcal{T}_{\text{RWTN}} \triangleq \langle \mathcal{K},
\hat{\mathcal{G}}_{RWFN} \rangle$ where a partial grounding
$\hat{\mathcal{G}}_{RWFN}$ can be described for predicates using
Eq.~(\ref{eq:rwfn}). Thus, we can easily compare the performance between
$\hat{\mathcal{G}}_{RWFN}$~(Eq.~(\ref{eq:rwfn})) and $\hat{\mathcal{G}}_{LTN}$~(Eq.~(\ref{eq:ltn_predicate})).

\subsection{Results}
\label{sec:results}

Our experiments mainly focus on the comparison of the performance
between our model and LTN, but figures also include results with
Fast-RCNN~\citep{girshick2015fast} for type
classification and the inclusion
ratio~$ir$ baseline in the
\emph{part-of} detection task. If $ir$ is greater than a given threshold
$th$~(in our experiments, $th = 0.7$), then the bounding boxes are said
to be in the \texttt{partOf} relation. Every bounding box $b$ is
classified into $C \in \mathcal{P}_1$ if $\mathcal{G}(C(b)) > th$.

\begin{figure*}[t!]\centering%
    \begin{subfigure}[t]{0.44\textwidth}\centering%
        \includegraphics[width=\textwidth]{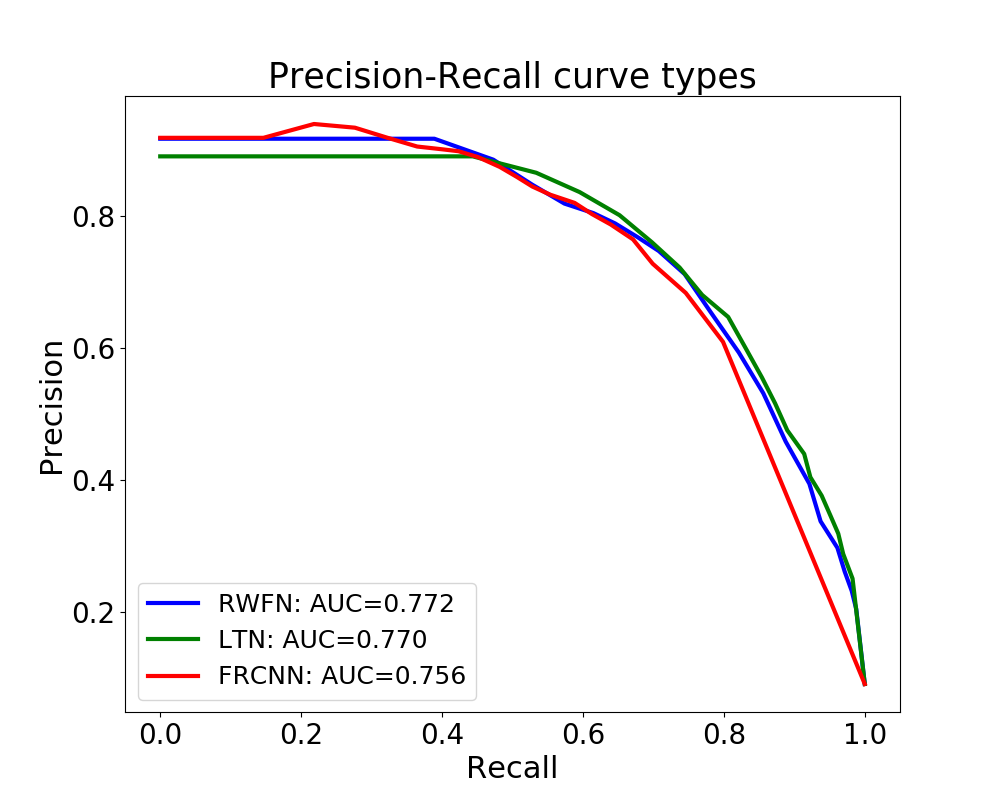}
        \caption{RWFNs achieve similar performance for object type
        classification compared to LTNs, achieving an Area Under the
        Curve~(AUC) of 0.772 (compared to 0.770).}
        \label{sii-indoor-result01}
    \end{subfigure}
    \hfill
    \begin{subfigure}[t]{0.44\textwidth}\centering%
        \includegraphics[width=\textwidth]{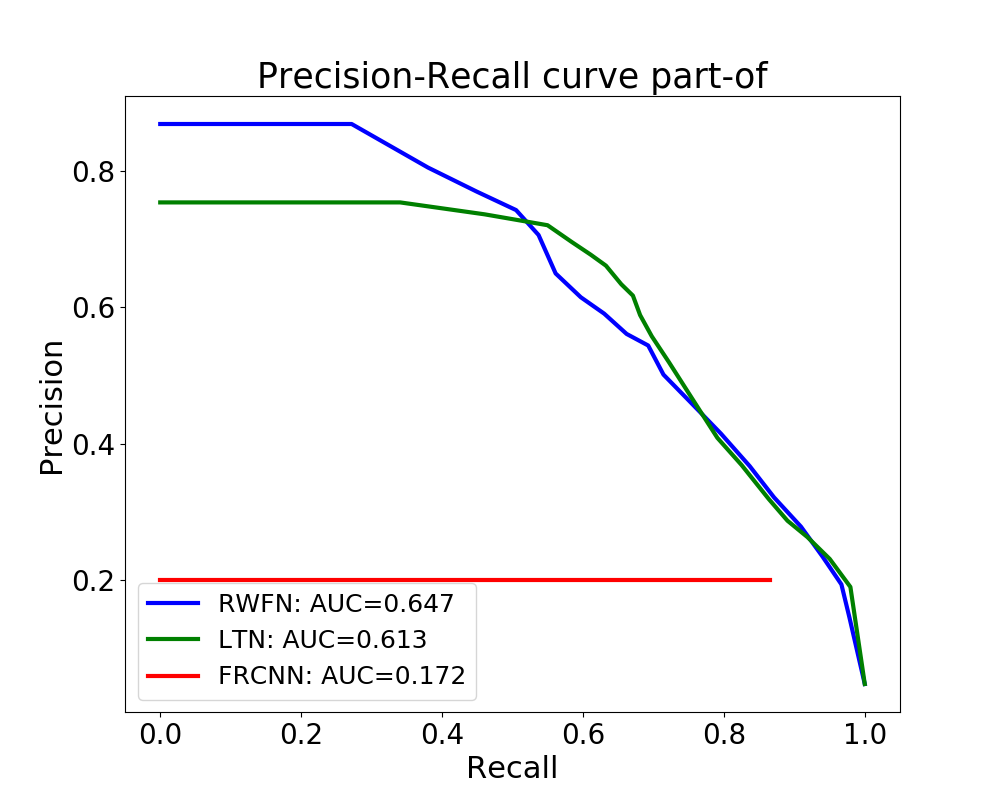}
        \caption{RWTNs outperform LTNs on the detection of \emph{part-of}
        relations, achieving AUC of 0.647 (compared to 0.613).}
        \label{sii-indoor-result02}
    \end{subfigure}
    \hfill
    \caption{Precision--recall curves for indoor objects type
    classification and the \texttt{partOf} relation between objects.}
    \label{sii-indoor-results}
\end{figure*}
Results for indoor objects are shown in Fig.~\ref{sii-indoor-results}
where AUC is the area under the precision--recall curve. The results
show that, for the \emph{part-of} relation and object types
classification, RWFNs achieve better performance than LTNs.
%
However, there is some variance in the results because of the
stochastic nature of the experiments. Consequently, we carried out five
such experiments for each task, for which the sample averages and 95\%
confidence intervals are shown in Table~\ref{table_performance}. These
results confirm that our model can achieve similar performance as LTNs
for object-task classification and superior performance for detection of
\emph{part-of} relations.

In Table~\ref{table_performance}, we only included AUC numbers for RWFNs
with weight sharing (third column) for object-type classification
because \emph{part-of} relations only require a single classifier. The
performance of RWFNs with weight sharing for the object-type
classification task (which requires 11 classifiers for indoor objects,
23 for vehicles, and 26 for animals) shows only a marginal gap in
performance compared to other models, which demonstrates the
effectiveness and efficiency of the approach of using a single shared encoder in RWFNs with weight sharing.
\begin{table}
    \caption{AUC of T1~(object type classification) and T2~(detection of \emph{part-of} relation) for LTN, RWFN, and RWFN with weight sharing across label groups. $\text{MEAN}_{\pm 2 \times \text{SD}}$ for all models. Best performances shown in \textbf{bold}.}
\label{table_performance}
\begin{tabular}{lccc}
\toprule
Label-Task & LTN & RWFN & RWFN w/ W.S \\
\midrule
Indoor-T1 & $.769_{\pm .0314}$ & $.770_{\pm .0092}$ & $\textbf{.773}_{\pm .028}$ \\
Indoor-T2 &  $.619_{\pm .082}$ & $\textbf{.648}_{\pm .0621}$ & --- \\
\hline
Vehicle-T1 & $.709_{\pm .0289}$ & $\textbf{.711}_{\pm .0162}$ & $.706_{\pm .0111}$ \\
Vehicle-T2 &  $.576_{\pm .0355}$& $\textbf{.613}_{\pm .0489}$ & --- \\
\hline
Animal-T1 & $\textbf{.701}_{\pm .024}$ & $.700_{\pm .024}$ & $.697_{\pm .0237}$ \\
Animal-T2 &  $.640_{\pm .0783}$& $\textbf{.661}_{\pm .0364}$ & --- \\
\bottomrule
\end{tabular}
\end{table}

As summarized in Appendix~\ref{app:ablation_study}, we also conducted ablation studies to assess the degree to which the AL--MB input transformation and the random Fourier features each contribute to the performance of the model. We have also included, in Appendix~\ref{app:performance_analysis}, a detailed comparison of performance among LTNs, RWFNs, and RWFNs with weight sharing. Specifically, we compare LTNs and RWFNs in terms of numbers of learnable parameters and running times; we also compare RWFNs with and without weight sharing in terms of space complexity.

\section{Conclusion and Future Work}
\label{sec:conclusion}

In this paper, we introduced Randomly Weighted Feature Networks, which
incorporate the insect-brain-inspired neuronal feature representation
and unique random features derived by random Fourier features. The RWFN
encoder acts as a generalized feature extractor with greater relational
expressiveness and a learning model with relatively simpler structure.
We demonstrated how insights from the insect nervous system can be
applied to the fields of neural-symbolic computing and knowledge
representation and reasoning for relational learning.

Our work can be advanced in several ways. For one, RWFNs can
be applied to other variants of SII problems proposed by
\citet{donadello2019compensating}, and performance between our model and LTN
for zero-shot learning in SII tasks can be compared. In addition, we plan to
extend application of RWFNs to tasks that need to extract structural knowledge
from not only images but also text, such as visual question-answering
challenges. Furthermore, we will investigate how other methods from
neuroscience for exploring biologically-plausible learning algorithms might be
applicable to our model.
Finally, we will extend RWFNs to include a recurrent part for
representing dynamic features of time-series data, similar to reservoir
computing~\citep{ferreira2009genetic, sun2017deep, wang2019echo}; this
approach may allow for extracting time-varying relational knowledge
necessary for developing a framework for data-driven reasoning over
temporal logic.

\begin{acknowledgments}
This work was supported in part by NSF SES-1735579.
\end{acknowledgments}

\bibliography{paper}

\appendix

\section{The Intuitions of RWFNs}
\label{app:intuition}
For the bio-inspired representation in our model, we
concentrated on implementing: (i) how to build random connections
between the AL glomeruli~(input layer) and the KCs~(hidden layer), and
(ii) how to guarantee hidden-layer sparsity to best differentiate one
odor stimulus from another. We used a sparse, binary, and random matrix
to define an arbitrary set of inputs for each KC in the model with
inspiration from \citet{caron2013random}, \citet{peng2017simple}, and
\citet{dasgupta2017neural}. In particular, in biological models of the
insect brain and the AL--MB interface, the firing rates from~7
randomly selected glomeruli are passed and summed to each
KC~\citep{caron2013random, inada2017origins}. Furthermore,
\citet{endo2020synthesis} developed a computational model of sparsity of
the KCs' output activity based on global inhibition the average KC
input. In their model, KCs output an intermediate result subject to
global inhibition from the average glomerular input to all KCs. The last
KC activity is then produced by thresholding the inhibited output
through a ramp function, which is functionally equivalent to the
Rectified Linear Unit~(ReLU) activation function~\citep{glorot2011deep}.
Thus, the output of the $j$th KC, $KC_{out_{j}}$, in the computational
model was described as:
\begin{equation}
    KC_{out_{j}}
    = \phi\left( KC_{in_{j}} - C\frac{1}{N_{KC}}\sum_{j}^{N_{KC}}KC_{in_{j}} \right)
\label{eq:r_out_bio}
\end{equation}
where $KC_{in_{j}}$ indicates the weighted sum of input from 7 random
indices of the input vector, $\phi$ is the ReLU, $C$ is the strength of
global inhibition, and $N_{KC}$ is the total number of KCs. The
parameters $C=1.0$ and $N_{KC}=2000$ were chosen so as to match the
values best calibrated to real KC responses~\citep{endo2020synthesis,
aso2009mushroom}. With this KC representation, \citet{endo2020synthesis}
trained a linear decoder to successfully classify 'group' from
'non-group' odors. Similarly, we make use of Eq.~(\ref{eq:r_out_bio}) and
train a linear model for learning latent relationships among input.

For another hidden representation using random Fourier features in our
model, based on Eq.~(\ref{eq:random_fourier_feature}), we can define a
decision function $f(\textbf{x})$ given a dataset including $N$ data
samples $\textbf{x}, \textbf{y} \in \mathbb{R}^{d}$ and a randomized
feature mapping $\textbf{z}: \mathbb{R}^{d} \to \mathbb{R}^{D}$ as
follows:
\begin{equation}
    \begin{split}
        f(\textbf{x}) &= \sum_{n=1}^{N} \alpha_{n} k(\textbf{x}_{n}, \textbf{x}) = \sum_{n=1}^{N} \alpha_{n} \langle \phi(\textbf{x}_{n}), \phi(\textbf{x}) \rangle \\
        &\approx \sum_{n=1}^{N} \alpha_{n} \textbf{z}(\textbf{x}_{n})^{\top} \textbf{z}(\textbf{x}) = \boldsymbol\beta^{\top} \textbf{z}(\textbf{x}) \\
    \end{split}
    \label{eq:rff_linear}
\end{equation}
This indicates that if $\textbf{z}(\cdot)$ can approximate $\phi(\cdot)$
well, we can simply map our data using $\textbf{z}(\cdot)$ and then use
a linear model to learn because both $\boldsymbol\beta$ and
$\textbf{z}(\cdot)$ in the above equation are $D$-vectors. Therefore,
the task that we will describe in the next section is how to find a
random projection function $\textbf{z}(\cdot)$ that can approximate the
corresponding nonlinear kernel machine appropriately.

The reason why we leverage the random Fourier feature is conciseness and
efficiency of computing linear interactions among input, which can be
replaced with the bilinear model in Eq.~(\ref{eq:ltn_predicate}). In
Eq.~(\ref{eq:ltn_predicate}), the bilinear tensor was used to compute
the relation, which seems intuitive because each slice of the tensor
serves as being responsible for one type of relation. However, this
computation requires high computational cost with large number of
parameters. In contrast, the random Fourier features in
Eq~(\ref{eq:rff_linear}) can do the similar task with a much faster
learning process and fewer number of parameters.

Considering how Eq.~(\ref{eq:r_out_bio}) and Eq.~(\ref{eq:rff_linear})
can be used in our model, the hidden representation can be expressed
with the concatenation of $KC_{out}$ and $\textbf{z}(\cdot)$. 

\section{Details of Experiments}
\label{app:details_exp}

\paragraph{Hardware specification of the server}
The hardware specification of the server that we used to experiment is as follows:
\begin{itemize}
    \item CPU: Intel\textregistered{} Core\textsuperscript{TM} i7-6950X CPU @ 3.00GHz (up to 3.50 GHz)
    \item RAM: 128 GB (DDR4 2400MHz)
    \item GPU: NVIDIA GeForce Titan Xp GP102 (Pascal architecture, 3840 CUDA Cores @ 1.6 GHz, 384 bit bus width, 12 GB GDDR G5X memory)
\end{itemize}

\paragraph{Source codes} 
All source codes, trained models, and figures in this paper are available at \url{https://github.com/jyhong0304/SII}.

\paragraph{Datasets}

The \textsc{PASCAL-Part}-dataset~\citep{chen2014detect} and
ontologies~(\textsc{WordNet}) are chosen for the \emph{part-of}
relation. The \textsc{PASCAL-Part}-dataset contains 10103 images with
bounding boxes. They are annotated with object-types and the part-of
relation defined between pairs of bounding boxes. There are three main
groups in labels---animals, vehicles, and indoor objects---with their
corresponding parts and ``part-of'' label. There are 59 labels~(20
labels for whole objects and 39 labels for parts).
The images were then split into a training set with 80\% of the images
and a test set with 20\% of the images, maintaining the same proportion
of the number of bounding boxes for each label. Given a set of bounding
boxes detected by an object detector~(Fast-RCNN), the task of object
classification is to assign to each bounding box an object type. The
task of \emph{part-of} detection is to decide, given two bounding boxes,
if the object contained in the first is a part of the object contained
in the second.

\paragraph{Hyperparameter Setting}

To compare the performance between RWFNs and LTNs, we trained two models
separately. For LTN, we configure the experimental environment
following~\citet{donadello2017logic}. The LTNs were configured with a
tensor of $k=6$ layers. For RWFN, the number of hidden nodes $B=200$ for
a classifier for object type classification. In addition,
we set the number of hidden nodes of a classifier for \emph{part-of}
detection as twice as large as the number of hidden nodes $B$ of a
classifier for object classification, which is 400. This is because the
dimension of inputs that the classifier for detecting the \emph{part-of}
relation is twice as large as the input space
required for a classifier for objection categorization.
%
Referring to \citet{donadello2017logic}, both models make use of a
regularization parameter $\lambda=10^{-10}$, Lukasiewicz’s
$t$-norm~($\mu(a, b) = \operatorname{max}(0, a + b - 1)$), and the
harmonic mean as an aggregation operator. We ran 1000 training epochs of
the RMSProp~\citep{Tieleman2012} learning algorithm available in
TensorFlow for each model.

\paragraph{Hyperparameter Searching for RWFNs}
To find out the best number of hidden nodes $B$, we used the Optuna framework~\citep{akiba2019optuna} with 500 iterations in the range of $[64, 512]$. The Optuna framework allows us to dynamically construct the parameter search space because we can formulate hyperparameter optimization as the maximization/minimization process of an objective function that takes a set of hyperparameters as input and returns a validation score. In our case, the validation score returned was the test AUC values. Furthermore, it provides efficient sampling methods, such as relational sampling that exploits the correlations among the parameters.

\section{Ablation Studies}
\label{app:ablation_study}

Table~\ref{tab:ablation} shows the results of ablation studies. In order to show how much two hidden representations~-- the input transformation between AL--MB and random Fourier features~-- contribute to the performance of our model, we built two separate RWFN models: one using the AL--MB input transformation only and another using random Fourier features only. Then, we performed five experiments and averaged AUCs of each model for object classification and \emph{part-of} detection. The number of hyperparameter $\beta$ for each model was set to the same as the number of hyperparameter for the original RWFN, which is 200.

For the object-type classification and \emph{part-of} detection tasks using Indoor label, the random Fourier features outperform the AL--MB input transformation. On the other hand, the AL--MB input transformation for \emph{part-of} detection tasks using Vehicle and Animal labels show better performance than the random Fourier features. Therefore, these ablation studies show that the model architecture of RWFNs in Eq.~(\ref{eq:rwfn}) can fully utilize both hidden representations and contribute to their good performance shown in Table.~\ref{table_performance} by compensating for each other.

\begin{table}
\caption{AUC of T1~(object type classification) and T2~(detection of \emph{part-of} relation) for the AL--MB representation~(AL--MB) and random Fourier features (RFF) across label groups. $\text{MEAN}_{\pm 2 \times \text{SD}}$ for all models. The best performance is displayed in bold.}
\label{tab:ablation}
\begin{tabular}{lccc}
\toprule
Label-Task & AL--MB & RFF  \\
\midrule
Indoor-T1 & $.743_{\pm .021}$ & $\textbf{.766}_{\pm .012}$ \\
Indoor-T2 &  $.525_{\pm .102}$ & $\textbf{.641}_{\pm .010}$ \\
\hline
Vehicle-T1 & $.710_{\pm .017}$ & $\textbf{.715}_{\pm .009}$ \\
Vehicle-T2 &  $\textbf{.612}_{\pm .027}$& $.572_{\pm .079}$ \\
\hline
Animal-T1 & $.705_{\pm .017}$ & $\textbf{.709}_{\pm .013}$ \\
Animal-T2 &  $\textbf{.664}_{\pm .069}$& $.646_{\pm .020}$ \\
\bottomrule
\end{tabular}
\end{table}

\section{Performance Analysis}
\label{app:performance_analysis}

\paragraph{Relative Complexity of RWFNs and LTNs}
To better appreciate the relative performance of RWFNs and LTNs, we can
compare the number of parameters for grounding a unary predicate for
each model. The dimension of the input in the dataset for both RWFNs and
LTNs is $n = 64$.
As shown in Eq.~(\ref{eq:ltn_predicate}), the
parameters to learn in LTNs are $\{ u_{P} \in \mathbb{R}^k,
W_{P}^{[1:k]} \in \mathbb{R}^{n \times n \times k}, V_{P} \in
\mathbb{R}^{k \times n}, b_P \in \mathbb{R}^k \}$, where $k = 6$
following the configuration of the LTNs. Thus, the number of parameters
in LTNs is $(n^2 + n + 2) \cdot k = (64^2 + 64 + 2) \cdot 6 = 24972$.
On the other hand, in Eq.~(\ref{eq:kcnet_intermediate}) and Eq.~(\ref{eq:rwfn_second_rep}), the number of parameters in
RWFNs are $\{ \textbf{W} \in \mathbb{R}^{n \times B}, \textbf{R} \in
\mathbb{R}^{n \times B}, \textbf{b} \in \mathbb{R}^{B}, \boldsymbol\beta
\in \mathbb{R}^{2B} \}$, where $B = 200$ following the configuration of
the RWFNs. Therefore, the number of parameters in RWFNs is $(2n + 3)
\cdot B = (2 \cdot 64 + 3) \cdot 200 = 26200$.
Although our method requires more space complexity compared to
LTNs~($26200 > 24972$), the parameters $\{ \textbf{W}, \textbf{R},
\textbf{b} \}$ in RWFNs are randomly drawn and fixed weights. Thus, it
is also necessary to compare the number of learnable parameters across
the two models.

All of the above parameters of LTNs must be adaptable, whereas the
parameters to learn in RWFNs for object type classification are only
$\boldsymbol\beta \in \mathbb{R}^{2B}$. Thus, the number of learnable
parameters is $400$, which is much smaller than that of LTNs. This means
that the ratio of the two numbers of parameters to learn is about
$400:24972 \approx 1:62$. Consequently, non-adaptable parameters in
RWFNs can have significant power to represent the latent relationship
among objects so that the model can efficiently extract relational
knowledge even though using fewer adaptable parameters. Furthermore, the
number of LTN parameters heavily depends on the number of features,
whereas RWFNs are independent of the number of features. In principle,
this could allow the learning process in our model to be accelerated if
the feature representation from the encoder model is pre-processed and
stored.

\paragraph{Running Time}
\begin{figure}[t!]\centering%
\includegraphics[width=0.5\columnwidth]{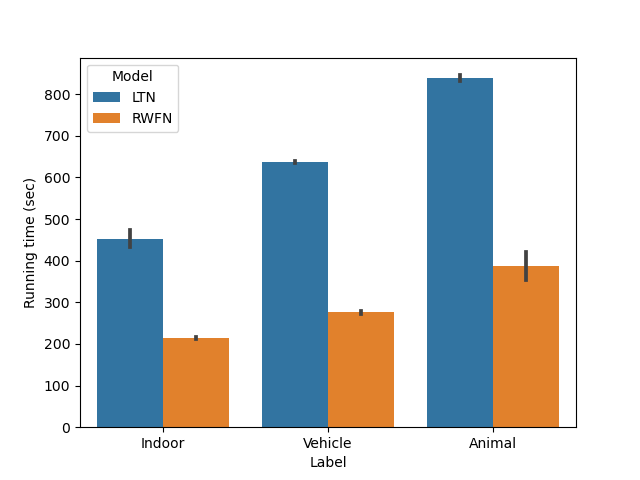}
\caption{The comparison of running time, including data configuration time, training time (sec) for LTN and RWFN}
\label{fig:running_time_models}
\end{figure}
Fig.~\ref{fig:running_time_models} depicts the comparison of running
time, including data configuration time and training time for LTNs and
RWFNs. The running time of RWFNs is roughly as half of that of LTNs.
This is because the number of learnable parameters in RWFNs is far
smaller than LTNs, and RWFNs have linear models to learn, which is much
simpler compared to models used in LTNs. 

\paragraph{Space Complexity of RWFNs with Weight Sharing}
Weight sharing is a unique feature of RWFNs, which can greatly reduce
necessary space complexity when multiple classifiers are used
simultaneously. In the depicted case of learning $i$ classifiers in
Fig.~\ref{fig:rwfn_ws_structure}, the space complexity for RWFNs without
using weight sharing is $(2 \cdot n \cdot B + 3 \cdot B) \cdot i = (2
\cdot n + 3)\cdot B \cdot i \approx O(i \cdot B \cdot n)$. However, with
weight sharing, RWFNs can achieve much better space complexity, which is
$2 \cdot n \cdot B + B + 2 \cdot B \cdot i \approx O(B \cdot n)$ because
$B \cdot i < B \cdot n$ for the experiments conducted in the SII task,
but it can also be $O(i \cdot B)$ for the different task. This indicates
that the one of the factors that highly influence to the space
complexity of the original RWFNs can be negligible when using weight
sharing, which makes the model cost efficient and economical.

\end{document}